\PassOptionsToPackage{dvipsnames,table}{xcolor}
\documentclass[10pt,twocolumn,letterpaper]{article}

\usepackage{cvpr}              %

\usepackage{booktabs, makecell, tabularx}
\usepackage{multirow}
\usepackage{duckuments}
\usepackage{amsmath}
\usepackage{graphicx}
\usepackage{arydshln}
\usepackage{booktabs}
\usepackage{amsmath}
\usepackage{listings}
\usepackage{todonotes}
\usepackage{algorithm}
\usepackage{algpseudocode}
\algrenewcommand\algorithmicrequire{\textbf{Inputs:}}
\algrenewcommand\algorithmicensure{\textbf{Outputs:}}

\definecolor{MyDarkBlue}{rgb}{0,0.08,1}
\definecolor{MyDarkGreen}{rgb}{0.02,0.6,0.02}
\definecolor{MyDarkRed}{rgb}{0.8,0.02,0.02}
\definecolor{MyDarkOrange}{rgb}{0.40,0.2,0.02}
\definecolor{MyPurple}{RGB}{111,0,255}
\definecolor{MyRed}{rgb}{1.0,0.0,0.0}
\definecolor{MyGold}{rgb}{0.75,0.6,0.12}
\definecolor{MyDarkgray}{rgb}{0.66, 0.66, 0.66}
\definecolor{MyDarkCyan}{rgb}{0.05, 0.55, 0.45}
\definecolor{MyBlack}{rgb}{0., 0., 0.}
\definecolor{MyMagenta}{rgb}{1., 0., 1.}
\definecolor{BkDarkBlue}{rgb}{.05,.07,.353}

\newcommand{\myparagraph}[1]{\noindent \textbf{#1} \ }

\newcommand{\ignorethis}[1]{}

\newcommand{\methodname}{\texttt{RefVFX}}
\usepackage{listings}
\usepackage{xcolor}
\usepackage{booktabs, tabularx, array}
\newcolumntype{Y}{>{\raggedright\arraybackslash}X}
\definecolor{cvprblue}{rgb}{0.21,0.49,0.74}
\usepackage[pagebackref,breaklinks,colorlinks,allcolors=cvprblue]{hyperref}

\usepackage{booktabs}     %
\usepackage{graphicx}     %

\def\paperID{4143} %
\def\confName{CVPR}
\def\confYear{2026}

\title{Tuning-free Visual Effect Transfer across Videos}

\author{
\makebox[\textwidth][c]{%
Maxwell Jones\textsuperscript{1} \quad
Rameen Abdal\textsuperscript{2} \quad
Or Patashnik\textsuperscript{2}
}
\\[0.5em]
\makebox[\textwidth][c]{%
Ruslan Salakhutdinov\textsuperscript{1} \quad
Sergey Tulyakov\textsuperscript{2} \quad
Jun-Yan Zhu\textsuperscript{1} \quad
Kuan-Chieh Jackson Wang\textsuperscript{2}
}
\\[0.75em]
\makebox[\textwidth][c]{%
\textsuperscript{1}Carnegie Mellon University \quad
\textsuperscript{2}Snap Research
}
}

\begin{document}

\twocolumn[{%
		\renewcommand\twocolumn[1][]{#1}%
		\maketitle
		\begin{center}
			\includegraphics[width=\linewidth]{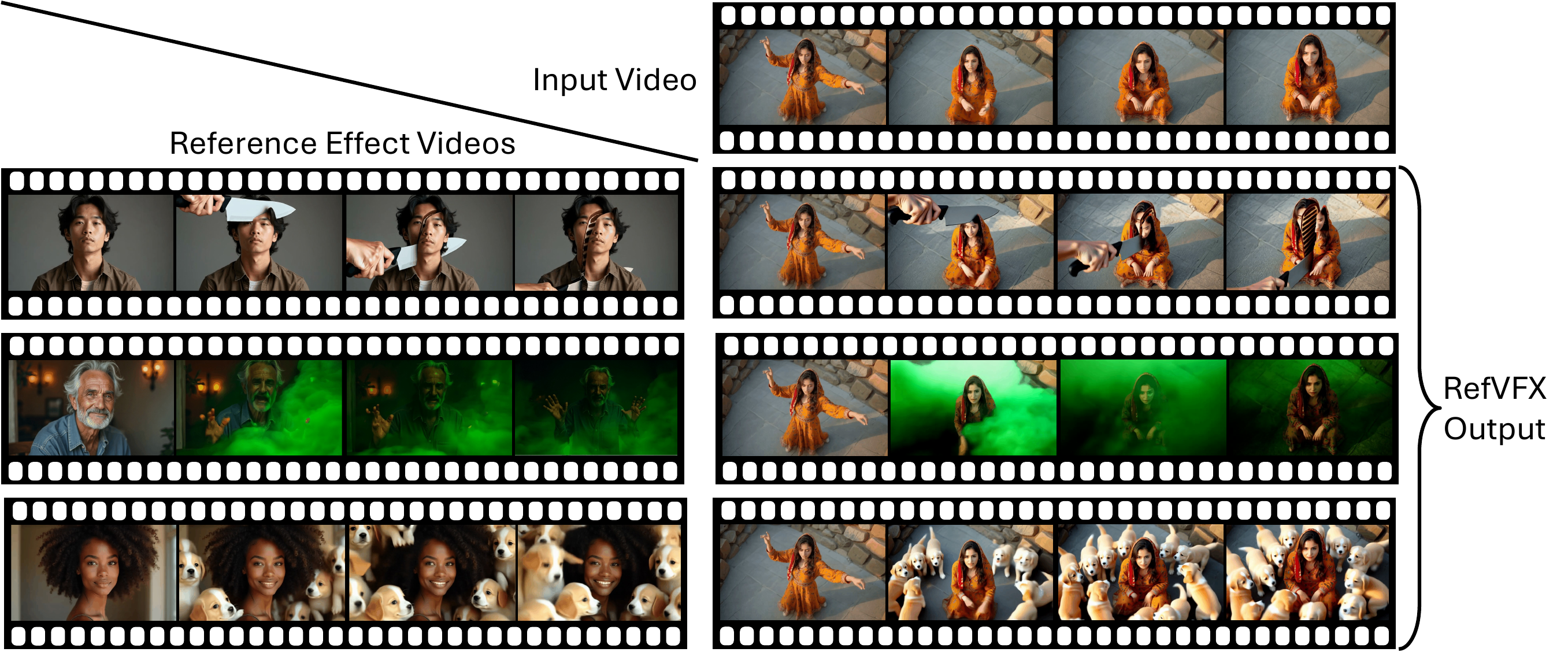}
            \vspace{-4mm} 
   \captionsetup{type=figure}
			\captionof{figure}{
               \textbf{Overview}. We present \methodname, a tuning-free framework for visual effect transfer across videos. Given a reference effect video and an input video, our method produces a new output video where the reference’s temporal effect is seamlessly applied to the input’s content and motion. Unlike prompt-based or keyframe-conditioned approaches, \methodname{} directly learns to interpret and reproduce complex time-varying visual effects such as material transformations, object additions, or atmospheric effects from example videos during training, enabling faithful and coherent visual effect transfer at inference time.}
               \vspace{2mm} 
			\label{fig:teaser}
		\end{center}
	}]

\begin{abstract}
We present \methodname, a new framework that transfers complex temporal effects from a reference video onto a target video or image in a feed-forward manner.
While existing methods excel at prompt-based or keyframe-conditioned editing, they struggle with dynamic temporal effects such as dynamic lighting changes or character transformations, which are difficult to describe via text or static conditions. 
Transferring a video effect is challenging, as the model must integrate the new temporal dynamics with the input video's existing motion and appearance. %
To address this, we introduce a large-scale dataset of triplets, where each triplet consists of a reference effect video, an input image or video, and a corresponding output video depicting the transferred effect.  Creating this data is non-trivial, especially the video-to-video effect triplets, which do not exist naturally. To generate these, we propose a scalable automated pipeline that creates high-quality paired videos designed to preserve the input's motion and structure while transforming it based on some fixed, repeatable effect. We then augment this data with image-to-video effects derived from LoRA adapters and code-based temporal effects generated through programmatic composition. 
Building on our new dataset, we train our reference-conditioned model using recent text-to-video backbones. 
Experimental results demonstrate that \methodname{} produces visually consistent and temporally coherent edits, generalizes across unseen effect categories, and outperforms prompt-only baselines in both quantitative metrics and human preference. See our website \href{https://snap-research.github.io/RefVFX/}{at this URL}.

\vspace{-15pt}
\end{abstract}
    
\vspace{-4pt}
\section{Introduction}
\label{sec:intro}
\vspace{-5pt}
Generative models have enabled the automatic synthesis and manipulation of images and videos with remarkable realism and diversity~\cite{blattmann2023stable, wan2025wan, kong2024hunyuanvideo, hacohen2024ltx, hong2022cogvideo, polyak2024movie, rombach2022high, podell2023sdxl, esser2024scaling, flux2024}. Recent advances in video generation have expanded user control, allowing video editing via text prompts, keyframes, or depth maps~\cite{wan2025wan, jiang2025vace, zhao2025controlvideo}. However, most models focus on semantic edits such as modifying objects~\cite{zi2025se, polyak2024movie}, scenes~\cite{zi2025se, polyak2024movie}, or styles~\cite{liu2023stylecrafter, ye2025stylemaster, zi2025se}, while overlooking \emph{``temporal effects''} (see Figure ~\ref{fig:teaser}): effects that \emph{unfold over time} and define the emotional, stylistic, and cinematic character of a video. Effects such as dynamic lighting, intricate camera movements, or character transformations remain difficult to express through text or current visual inputs.
 
Reference-based conditioning offers a natural way to specify complex temporal behaviors, letting users convey the desired effect through a reference video that captures subtle cues such as motion rhythm, lighting changes, or stylistic transitions. Transferring such temporal effects, especially from one video to another, i.e., a Ref.\ Video + Input Video~$\rightarrow$~Output Video setup, is especially challenging. To the best of our knowledge, we are the first to show results on this task.

While reference-based editing has shown promise for images~\cite{ye2023ip,kumari2023multi,ruiz2023dreambooth}, extending it to videos poses unique challenges. First, building a dataset is non-trivial: image-to-video (I2V) effects require consistent triplets in which a single static input produces multiple effect-rich outputs, and video-to-video (V2V) effects demand motion-consistent transformations that change only the temporal effect. Without such data, existing systems may fail to learn how to transfer temporal effects independently of the reference video’s content or motion. Second, the model must extract temporal dynamics from the reference and seamlessly integrate them with the target’s motion and appearance to produce coherent, high quality results. This requires disentangling the effect itself from the specific content and motion of the reference.

We introduce a reference-based video generation dataset, training method, and final framework of finetuned model along with an updated conditioning mechanism dubbed \methodname{}  that transfers complex temporal effects from a reference video onto a target image or video in a feed-forward manner. Built upon recent text-to-video diffusion backbones~\cite{wan2025wan, hacohen2024ltx, kong2024hunyuanvideo}, our model jointly conditions on three inputs: a reference video that provides the temporal effect, an input image or video that defines the scene content and motion, and a text prompt that offers high-level semantic guidance. Through this conditioning, \methodname{} learns to integrate the reference’s temporal dynamics harmoniously with the input’s appearance and motion, producing coherent and visually consistent results.

A key contribution of our work is a large-scale dataset designed specifically for this task. Each sample consists of a triplet: a reference video depicting a temporal effect, an input image or video, and a corresponding output video showing the effect applied to the input. Building such a dataset at scale is challenging, as it requires coherent alignment between the reference effect and the target’s content and motion. To this end, we construct a unified data curation pipeline that combines three complementary sources: (1) image-to-video effect data derived from existing LoRA-based models, (2) video-to-video transformations generated through an automated pipeline, and (3) synthetic, code-based temporal effects. Together, these sources yield over 1,700 unique effects and more than 120K triplets, providing unprecedented diversity for training reference-conditioned video editors.

We evaluate \methodname{} across diverse unseen temporal effects, comparing it with recent prompt-based and reference-guided video editing models~\cite{wan2025wan, jiang2025vace, Hleihil2025DecartAI}. Qualitatively, our method produces coherent, visually consistent videos that successfully integrate the transferred effects while preserving the motion and appearance of the input. Quantitatively, \methodname{} achieves higher scores in reference video embedding similarity to the reference effect video; however, existing metrics only partially reflect the aesthetic and temporal nuances of effect transfer. To better assess perceptual quality, we conduct a human preference study, where participants consistently favor our results, highlighting the importance of human judgment for evaluating dynamic visual effects. Notably, the model generalizes to unseen categories of effects and operates in a feed-forward manner without optimization at inference time, demonstrating both robustness and efficiency. Our code will be released upon publication. 

\noindent Our main contributions are summarized as follows:
\begin{itemize}
\item We introduce \methodname{}, a framework for reference-based video effect transfer that enables users to apply complex temporal effects from a reference video onto arbitrary videos or images in a feed-forward manner. 
\item We construct a large-scale, effect-aligned dataset comprising over 120K triplets of (reference, input, output) videos covering more than 1,700 distinct temporal effects along with a validation set, establishing a new benchmark for future research. %
\item We design a multi-source conditioning architecture built upon recent diffusion backbones that jointly encode reference video dynamics, input appearance/motion, and text prompts. %
\item We conduct extensive qualitative, quantitative, and human preference evaluations, demonstrating that our method achieves superior visual and temporal coherence compared to prompt-only or reference-guided baselines, and generalizes effectively to unseen effect categories.
\end{itemize}

\section{Related Work}
\label{sec:related_works}

\myparagraph{Text-to-Video Generation.}
Recent advances in text-to-video generation have significantly improved video quality, temporal coherence, and prompt following~\cite{blattmann2023stable, wan2025wan, kong2024hunyuanvideo, hacohen2024ltx, hong2022cogvideo, polyak2024movie, ma2024latte, cloneofsimo2024auraflow, ho2022imagen, ho2022video}. 
Similar to text-to-image generation~\cite{rombach2022high, podell2023sdxl, esser2024scaling, flux2024}, these models are largely based on diffusion or flow-matching in latent space~\cite{song2020denoising, dhariwal2021diffusion, lipman2022flow, albergo2023stochastic, liu2022flow}. 
Early work adopted U-Net backbones~\cite{ronneberger2015u, singer2022make, guo2023animatediff}, while modern approaches employ diffusion transformers~\cite{peebles2023scalable} that embed videos into latent patches and apply bidirectional attention with text~\cite{wan2025wan, hacohen2024ltx, polyak2024movie, ho2022imagen, hong2022cogvideo, kong2024hunyuanvideo}. 
Beyond pure text conditioning, some models perform image-to-video or first–last-frame generation~\cite{wan2025wan, hacohen2024ltx}, where a user provides keyframes and a text prompt to guide synthesis. 
However, these settings modify only edge frames, whereas our approach conditions on both an entire input video and a reference effect video, enabling transfer of \emph{temporal effects} that evolve throughout the sequence while preserving input motion and appearance.

\myparagraph{Reference-Based Controllable Generation.}
Reference-based conditioning uses a reference image or video that is not spatially aligned with the output. 
In text-to-image models, it is widely used for identity preservation~\cite{kumari2025generating, chen2024anydoor, li2023blip, parmar2025object, song2024moma, wei2023elite, xiao2025fastcomposer, guo2024pulid, ye2023ip, patashnik2025nested, ruiz2023dreambooth} or style transfer~\cite{huang2023composer, hertz2024style, rout2024rb, gao2025styleshot, xing2024csgo, zhang2024finestyle, sohn2023styledrop}. 
These signals are injected via per-reference optimization~\cite{sohn2023styledrop, gao2025styleshot, ruiz2023dreambooth}, cross-attention~\cite{li2023blip, parmar2025object, guo2024pulid, ye2023ip, song2024moma, xing2024csgo}, or added reference tokens~\cite{kumari2025generating, hertz2024style, rout2024rb}. 
Video generative models extend these paradigms temporally~\cite{wang2023videocomposer, liu2025phantom, liu2023stylecrafter, he2024id, yuan2025identity, huang2025conceptmaster, liang2025movie, chen2024videoalchemy, ye2025stylemaster}, typically conditioning on one or more \emph{static} reference images (identity or style) alongside text. 
Such methods maintain consistent appearance but cannot model \emph{temporally evolving} effects. 
More recent finetuning based methods such as Dynamic Concepts~\cite{dynamic_concepts, abdal2025zeroshotdynamicconceptpersonalization} are also inefficient, requiring either a new LoRA per effect instance or trained on limited data hampering scalability. 
In contrast, \methodname{} conditions on a \emph{reference video} that encodes dynamic, time-varying effects, allowing generalizable transfer of phenomena such as lighting changes, camera motion, stylistic transitions or identity-based conditioning.

\myparagraph{Video Editing.}
Video editing has progressed through both zero-shot and supervised methods~\cite{brooks2023instructpix2pix, labs2025flux, wu2025qwen, wei2024omniedit, hertz2022prompt, tumanyan2023plug, cao2023masactrl, zhang2023magicbrush, kulikov2025flowedit, brack2024ledits++, deutch2024turboedit}. 
Pretrained text-to-image models can perform zero-shot edits~\cite{hertz2022prompt, tumanyan2023plug, cao2023masactrl, kulikov2025flowedit, brack2024ledits++, deutch2024turboedit}, while paired datasets of object or style edits~\cite{wei2024omniedit, boesel2024improving} have enabled general-purpose editing models~\cite{brooks2023instructpix2pix, labs2025flux, wu2025qwen}. 
Video editing methods follow similar patterns: some are zero-shot~\cite{ceylan2023pix2video, geyer2023tokenflow}, while others train paired text-driven editors~\cite{cheng2023consistent, zi2025se, liu2025generative}. 
However, existing methods rely solely on text and an input video, enabling semantic or appearance edits but offering no mechanism to control \emph{how} visual properties evolve over time. 
Our approach introduces an additional reference video encoding a temporally varying effect, enabling control over dynamic phenomena such as lighting transitions, motion-dependent effects, and stylistic evolution that static, per-frame, or text-only paradigms cannot capture.

\vspace{-5pt}
\begin{figure*}[t!]
    \centering
    \includegraphics[width=\linewidth]{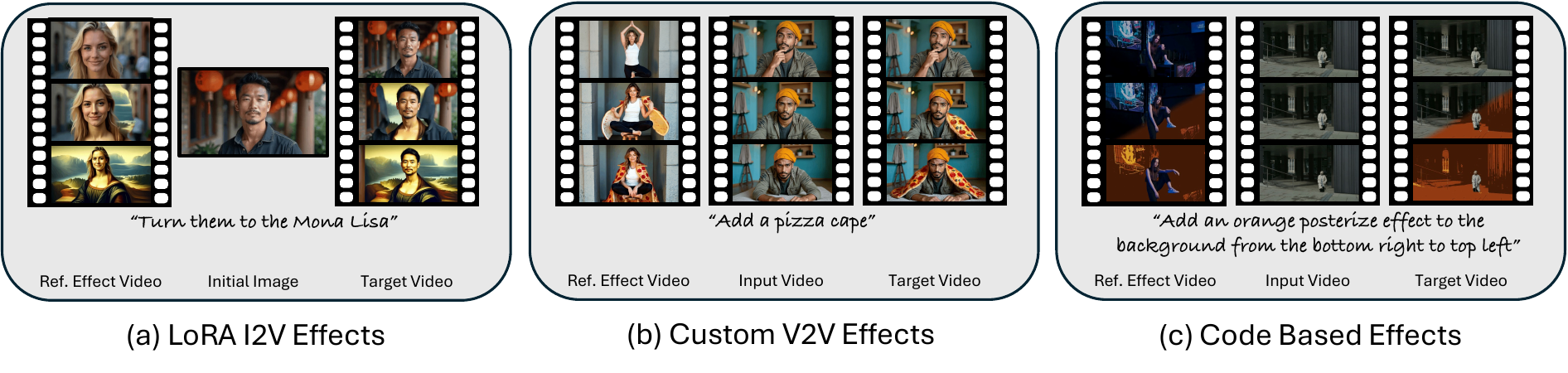}
        \vspace{-15pt}
    \caption{\textbf{Dataset Examples}. We show example triplets from each of our datasets. (a) We curate a reference video + image to video dataset by collecting Low Rank Adapters (LoRAs) \cite{hu2022lora} for different Image to Video effects online. For each effect, we can apply its corresponding LoRA to two separate images to create a triplet. (b) We create a custom pipeline for generating text guided reference video + input video to video effects. For more details, see Figure \ref{fig:v2v_method}. (c) We generate a large scale set of (ref video, input video, output video) triplets by curating specific code pipelines that apply specific, detailed effects to arbitrary videos. Armed with a specific code based effect and a fixed set of hyperparameters, we can apply the exact effect to an arbitrary number of input videos to create many triplets.}
    \label{fig:dataset}
    \vspace{-15pt}
\end{figure*}

\section{Method}
\label{sec:method}

\subsection{Overview}
\label{sec:overview}
Our goal is to transfer a temporally evolving visual effect from a \emph{reference video} onto a separate \emph{input image or video}. The resulting output should preserve the content and motion of the input while exhibiting the temporal dynamics shown in the reference. To achieve this, we introduce \methodname{}, which combines (1) a large-scale effect–aligned triplet training dataset, (2) a pretrained diffusion-based video generation model, and (3) an updated conditioning scheme allowing the video generation model to condition jointly on the reference video, input video, and text.

First, we give a high level overview of our dataset, followed by describing how we create each individual subset of the dataset in detail. Following this, we describe the model architecture for \methodname{} and implementation details for training.

\subsection{Training Data}

\begin{figure}[t!]
    \centering
    \includegraphics[width=.8\linewidth]{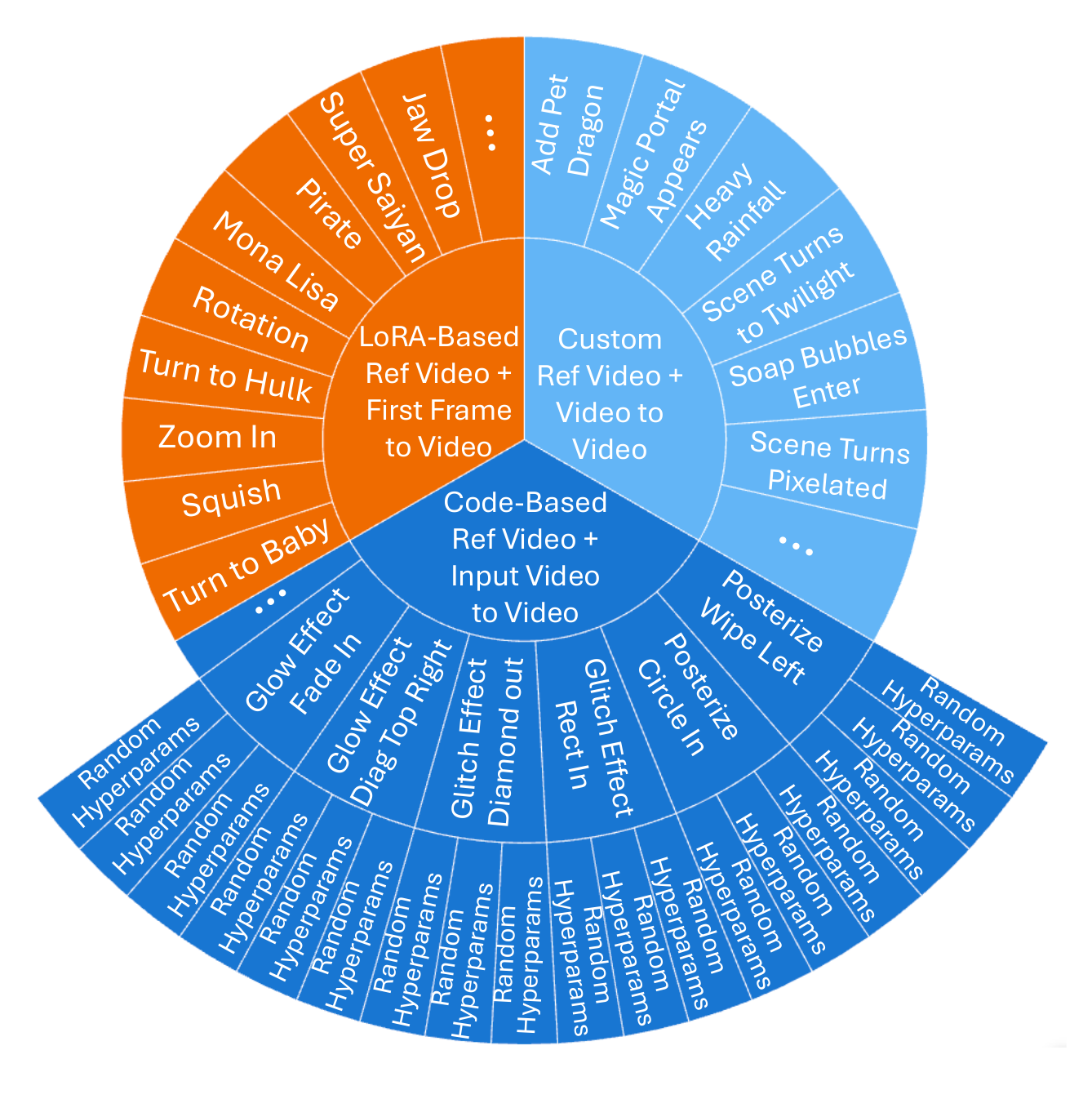}
        \vspace{-8pt}
    \caption{\textbf{Dataset Subset}. 
    We display a summary of our dataset structure and available sample effects. These include LoRA-based Image-to-Video, our scalable V2V pipeline (see Fig. 4) , and programmatic methods. For code-based effects, individual tasks are created by combining specific effects (e.g., Glow) and transitions (e.g., Fade In) with randomized hyperparameters.}
    \label{fig:dataset_subset}
    \vspace{-15pt}
\end{figure}

Our training data consists of triplets of the form $(\text{reference video},~\text{input image or video},~\text{target video})$. 
To create these triplets, we first curate $N$ effects $\{E^i\}_{i=1}^N$ (e.g., rain, motion blur, color shift), with each effect $E^i$ associated with a set $S^i$ of $K$ input output pairs: $S^i=\{(\text{input}^i_j,\text{output}^i_j)\}_{j=1}^K$. Here, $\text{output}^i_j$ is obtained by applying $E^i$ to $\text{input}^i_j$. 
To form a training triplet, we select an effect $E^i$ and two distinct pairs $(\text{input}^i_j, \text{output}^i_j)$ and $(\text{input}^i_{\ell}, \text{output}^i_{\ell})$. We discard the input of the first pair, and construct triplet $(\text{reference}=\text{output}^i_j,~\text{input}=\text{input}^i_{\ell},~\text{target}=\text{output}^i_{\ell})$, where the model takes in the first two videos and must predict the third. 
With this objective, the model learns to transfer the temporal effect from the reference onto the input while preserving the input videos motion and content. 
Triplets are derived from three complementary sources: (1) LoRA-based image to video effects, (2) video to video effects generated through our scalable pipeline, and (3) programmatically defined temporal effects, which we describe below and showcase in Fig.~\ref{fig:dataset}.

\subsubsection{Image to Video Effects:}

We curate a set of image-to-video effects using open-source Low-Rank Adapters (LoRAs)~\cite{hu2022lora} trained on small sets of videos sharing a common visual effect atop a base image-to-video model~\cite{wan2025wan}. 
Each adapter is treated as a distinct effect $E_i$. For every effect, we generate a large collection of videos from diverse input images synthesized via high-quality image generation models~\cite{wu2025qwen, labs2025flux}, where each per-adapter set forms one effect category. 
Triplets are composed of a reference video with effect $E_i$, the first frame of the target, and the corresponding output video. 
In total, we source 43 LoRAs with an average of 300 videos per effect, yielding over 14K video clips. See Fig. \ref{fig:dataset_subset} for example LoRAs.

\subsubsection{Video to Video Effects:}
\label{sec:v2v_algo}
We propose a scalable video-to-video data generation pipeline for a broad class of \textit{motion-based effects}. These effects manifest \textit{over time} rather than being the product of a single frame image edit followed by propagation of the edit to all frames (e.g., "turn the person into metal" as compared to "add a hat to the person throughout"). To our knowledge, this is the first scalable method designed for creating a large-scale training data pipeline for motion-based video editing.

\begin{figure*}[t!]
    \centering
    \includegraphics[width=\linewidth]{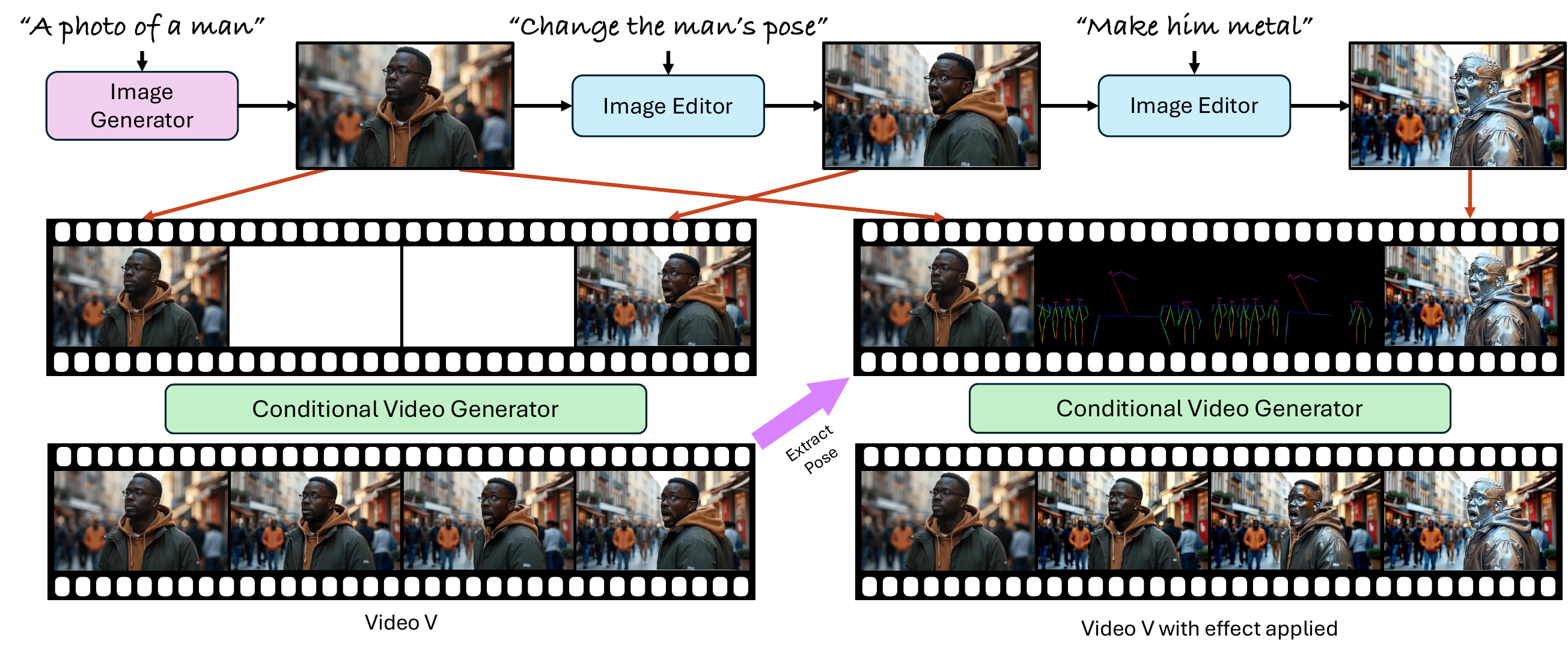}
        \vspace{-5pt}
    \caption{\textbf{Text-Guided Video Editing Pair Creation}. We present a method to generate a pair of videos $(V, V')$ from an effect prompt $E$, where $V$ is an initial video and $V'$ is the video with an effect $E$ applied. First, an image generation model is used to create an initial image. Next, an image editing model is used to change the pose, camera angle, and facial expression of the image. Finally, this image is again edited to add effect $E$. The first and second generated images are used with a first last frame model to output video $V$. Then, we use a conditional video model conditioned on the original first frame, effect edited last frame, \textit{and intermediate poses from video $V$} as conditioning to create video $V'$.}
    \label{fig:v2v_method}
    \vspace{-15pt}
\end{figure*}

Given an arbitrary effect prompt, our method produces paired videos $(V,V')$, consisting of an original video $V$ and its edited counterpart $V'$ exhibiting effect $E_i$ (Fig.~\ref{fig:v2v_method}, \ref{fig:dataset_subset}).
\myparagraph{Scalable V2V Algorithm.}
The process for generating these $(V, V')$ pairs is as follows:
We begin by generating a subject image $I$ using a high-quality image generation model \cite{flux2024, wu2025qwen}.
Using an image editing model \cite{labs2025flux, wu2025qwen}, we modify $I$ to produce an image $I'$ that depicts the same subject under a new pose, camera angle, and facial expression, which will be used to anchor the last frame of our video pair.
To obtain the last frame of the edited video with effect $E_i$, we apply an image editing model on $I'$ \cite{labs2025flux, wu2025qwen} to apply effect $E_i$, resulting in $I''$.  
Given the unedited image $I$ and the edited image $I'$, we synthesize video $V$ using a first–last frame interpolation model \cite{wan2025wan}. Once we have $V$, we extract intermediate poses using an image-to-pose model \cite{yang2023effective}, which will be used as conditioning for our second video generation.
Finally, we generate the edited video $V'$ by leveraging a conditional video generation model \cite{jiang2025vace}, which can take in independent conditioning signals for each output video frame generated. We take advantage of this property, and condition the video generation on:  
   \begin{itemize}
       \item [-] Initial image $I$ for first frame conditioning
       \item [-] Final edited image $I''$ for last frame conditioning
       \item [-] Intermediate pose images extracted from $V$ for frames 2 through $N - 1$
   \end{itemize}
   This enables framewise consistency between the input and output video as well as realistic temporal transitions between the unedited initial frame $I$ and stylized final frame $I''$. For a visual, see Figure \ref{fig:v2v_method}.

\vspace{-17pt}

\paragraph{Effect Generation.}
To ensure diversity, we automatically generate a large set of motion-based effect prompts using GPT-4o \cite{hurst2024gpt}. These effects span multiple semantic and visual categories, which we detail in the supplement.

\subsubsection{Programmatic Temporal Effects.}
\label{sec:syth_data_curation}

We further construct a large-scale dataset of \textit{programmatically defined temporal effects} applied to real videos. Our approach enables scalable and reproducible creation of diverse video effects with fine-grained control over appearance and temporal dynamics.

\myparagraph{Base Videos.}
We source input videos from the Senorita dataset \cite{zi2025se}, which provides high-quality clips with accompanying foreground–background segmentation masks obtained using SegmentAnything \cite{ravi2024sam}. These masks allow effects to be applied selectively to the full frame, the foreground, or the background.

\myparagraph{Effect Library.}
We define a library of code-based effects ${c_i}$ such as posterization, pixelation, and dithering. Each effect includes hyperparameters that control its visual characteristics. For example, posterization varies in the number of color bins and palette type, while pixelation varies in block size. Sampling different parameter configurations yields a wide range of distinct appearances within each effect class.

\myparagraph{Temporal Compositing.}
To introduce temporal variation, we consider a set of temporal transition operators $t_i$ (e.g., wipe-right, diagonal wipe, circular-out, etc). Each transition is also parameterized by temporal hyperparameters controlling its start time and duration.

\myparagraph{Final Composition.}
For each composite effect $E_i$, we randomly sample an effect mask (e.g., foreground, background, or all), a spatial effect, its parameters, a temporal transition, and its corresponding temporal parameters. See Figure \ref{fig:dataset} (c) for an example.
This final effect $E_i$ is then applied to a fixed set of source videos, generating $K$ unique clips per effect. In total, this pipeline yields approximately 100K videos spanning 1,500 distinct synthetic motion-based effects.
\vspace{-5pt}
\subsection{Model Architecture and Conditioning}
\vspace{-1pt}
\begin{figure*}[t!]
    \centering
    \includegraphics[width=.95\linewidth]{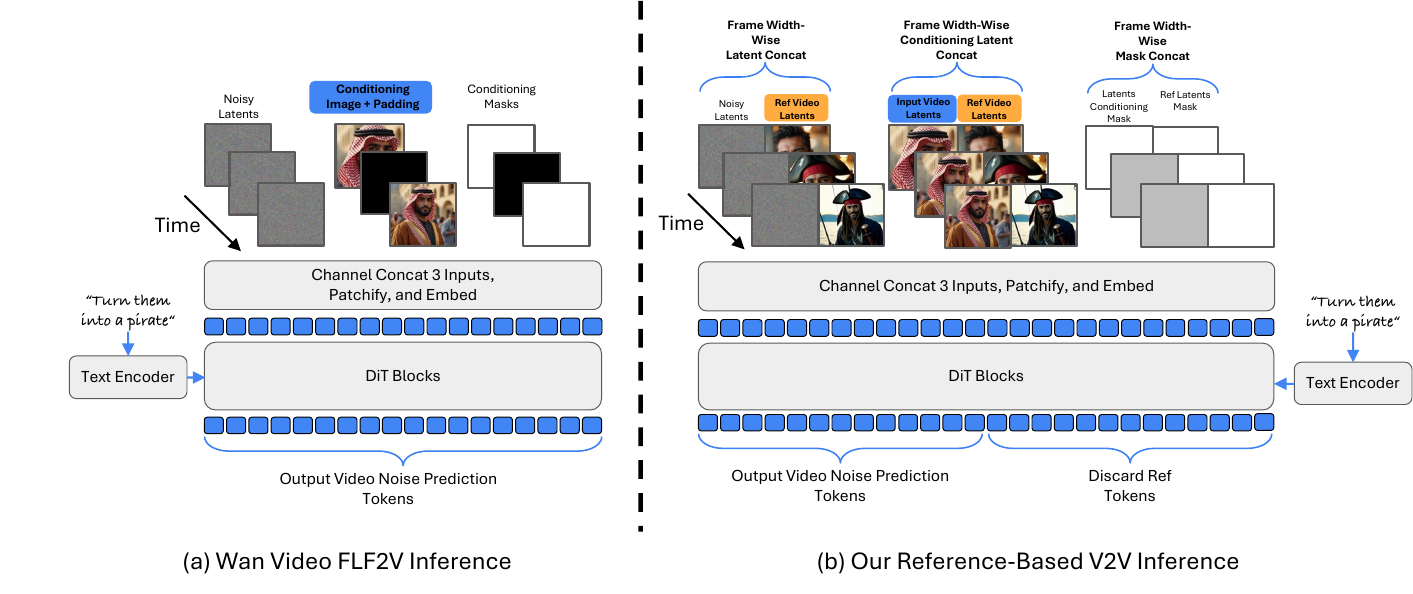}
        \vspace{-5pt}
    \caption{\textbf{Architecture overview}. (a) Standard Wan Video First–Last Frame to Video (FLF2V) architecture: noisy spatio-temporal latents are channel-wise concatenated with conditioning inputs and a mask, then patchified, embedded, and processed by the diffusion transformer to predict velocity.
(b) In our setup, x \textcolor{blue}{input video latents} are used as conditioning for the noisy latents, while \textcolor{orange}{reference video  latents} are concatenated width-wise to both. The latent mask is set to 1 for frames preserved exactly in the output and 0.5 for those to be modified; the reference latent mask is all ones. This design doubles the token count relative to base generation while conditioning on both the reference effect video and input video. Since all three inputs are channel-concatenated before patchification, repeated clean reference latents are merged channel-wise before embedding, ensuring no redundant reference information across tokens.}
    \label{fig:architecture}
    \vspace{-15pt}
\end{figure*}

We build upon the Wan2.1 conditioning model architecture \cite{wan2025wan, jiang2025vace}, which extends text-to-video diffusion models to image-to-video (I2V), first-last frame to video (FLF2V), depth to video, and other conditioning types by concatenating noisy latent channels with conditional and mask channels. We extend this mechanism to jointly condition on both a reference video effect and an input video.

The input video’s latents are supplied as conditioning latents, while the reference effect video is encoded into additional latents that are concatenated across all frames. This allows for spatial self-attention between noisy latent tokens with input video conditioning information and reference effect tokens during a forward pass. A hybrid mask controls which latent conditioning frames are preserved or edited based on which are directly copied to the final output and are changeable. The architecture is shown in Figure \ref{fig:architecture}. Further architectural details, including input condition dropout and cross attention configurations are provided in the supplementary material.

\vspace{-5pt}
\subsection{Implementation Details}
We fine-tune the Wan2.1 \cite{wan2025wan} 14 Billion parameter First-Last Frame to Video model with Low Rank Adapters for 10K steps with a batch size of 8 on a single node of 8 NVIDIA A100 GPUs. We sample from each of the three datasets equally during training, and drop the reference effect video conditioning, control video conditioning, and text prompts with low probability to allow for various forms of classifier-free guidance \cite{ho2022classifier, brooks2023instructpix2pix} during inference time. We also add the true last frame from the target video to the model with low probability to allow for the first and last frame capabilities of the underlying model to be retained under our new setup.

\begin{table}[t]
\centering

\resizebox{\columnwidth}{!}{

\begin{tabular}{l|ccc|ccc}
\toprule
 & \multicolumn{3}{c|}{\textbf{Neural V2V}} & \multicolumn{3}{c}{\textbf{Code-Based V2V}} \\
\cmidrule(lr){2-4} \cmidrule(lr){5-7}
Baseline & RVA & IVA & OM & RVA & IVA & OM \\
\midrule
Lucy Edit & 62.3 $\pm$ 2.9 & 60.7 $\pm$ 2.8 & 65.7 $\pm$ 3.1 & 64.7 $\pm$ 4.5 & 61.4 $\pm$ 5.0 & 65.3 $\pm$ 5.0 \\
VACE (Depth) & 58.3 $\pm$ 2.5 & 59.1 $\pm$ 2.6 & 66.8 $\pm$ 2.1 & 65.7 $\pm$ 5.4 & 56.2 $\pm$ 5.0 & 61.9 $\pm$ 5.0 \\
VACE (Pose) & 57.0 $\pm$ 2.1 & 53.7 $\pm$ 2.2 & 60.0 $\pm$ 2.2 & 62.9 $\pm$ 4.8 & 63.3 $\pm$ 4.9 & 64.3 $\pm$ 4.7 \\
No Ref & 57.5 $\pm$ 2.8 & 51.3 $\pm$ 3.1 & 67.2 $\pm$ 2.9 & 59.5 $\pm$ 3.5 & 54.3 $\pm$ 4.0 & 61.3 $\pm$ 3.9 \\

\bottomrule
\end{tabular}

}
\\[0.5em]
\caption{User Study Results for Neural V2V and Code-Based Edits. Each cell shows Win Rate (\%) with standard deviation. Our method shows large preference over base models in Reference Video Adherence (RVA), Input Video Adherence (IVA), and Overall Match (OM), where we ask users to judge which method best adheres to the input video \textit{while applying the reference video effect.}}
\label{tab:v2v_code_user_study_detailed}
\vspace{-13pt}
\end{table}

\begin{table}[t]
\centering

\resizebox{.65\linewidth}{!}{

\begin{tabular}{lcc}
\toprule
Baseline & Reference Video Adherence & Overall Match \\
\midrule
Wan2.1 & 57.1 $\pm$ 2.4 & 63.4 $\pm$ 2.4 \\
VACE (I2V) & 57.4 $\pm$ 3.2 & 62.9 $\pm$ 2.6 \\
No Ref & 56.8 $\pm$ 2.3 & 61.8 $\pm$ 3.4 \\
\bottomrule
\end{tabular}

}
\caption{User Study Results for I2V. Each cell shows Win Rate (\%) with standard deviation. Our method shows large user preference over baselines in Reference Video Adherence and Overall Match (OM), where we ask users to judge which method best preserves the input image \textit{while applying the reference video effect.}}
\label{tab:i2v_user_study_detailed}
\vspace{-10pt}
\end{table}

\section{Experiments}
\label{sec:experiments}

We evaluate our approach on both reference-based image-to-video and reference-based video-to-video generation tasks. In each setting, we provide a pair 
(reference video, input image or video), and assess the quality of our generated outputs against strong baselines. Our evaluation emphasizes generalization to unseen reference effects and motion patterns.
\vspace{-5pt}
\subsection{Validation Datasets}

\myparagraph{Image-to-Video Testing.}
We construct an evaluation set of unseen image-to-video effects sourced from publicly available LoRA-based visual effect models that were not included in training. For each LoRA effect, we generate multiple reference videos from distinct input images, resulting in 28 unique unseen LoRA effects and a total of 56 reference videos for evaluation.

\myparagraph{Video-to-Video Testing.}
We generate a validation set of Reference Video Effect + Input Video pairs for our reference effect based video to video task. For the first source of data, we reuse the unseen LoRA effects from the previous section as reference effect videos and replace the input images with unseen real input videos to produce the input video pair. Secondly, we use our pipeline introduced in Section~\ref{sec:v2v_algo} with novel prompts not seen during training for a portion of the validation dataset. Each resulting pair 
(reference video, input video) generated using our method is manually filtered to ensure high perceptual quality. Finally, we synthesize an additional set of temporally varying effects using the programmatic procedure from Section~\ref{sec:syth_data_curation}, but with hyperparameter choices unseen during training. We collect over 100 validation instances, and separate the Code-Based Video-to-Video Effects from the Neural Video-to-Video effects when reporting results.

\subsection{Baselines}

Since existing baselines cannot directly condition on reference videos, we compare our method to state-of-the-art text-conditioned image-to-video and video-to-video models. All baselines receive the same textual descriptions of the desired effects as those implicit in our reference videos.

\myparagraph{Image-to-Video Baselines.}
We benchmark against the Wan 2.1 image-to-video model~\cite{wan2025wan} and the Wan VACE model~\cite{jiang2025vace}, which represent current state-of-the-art text-guided video generation systems.

\myparagraph{Video-to-Video Baselines.}
For the video-to-video task, we evaluate two configurations based upon the Wan VACE \cite{jiang2025vace} conditional model and one native text-based video editing model. Pose-conditioned Wan VACE: In this setup, we condition the Wan VACE on the first frame of the input video, intermediate poses extracted from the input video~\cite{yang2023effective}, and the corresponding text prompt. Depth-conditioned: In this setup, we condition the Wan VACE on the first frame of the input video, intermediate depth maps~\cite{ranftl2020towards} from the input video, and the same text prompt. Lucy Edit \cite{Hleihil2025DecartAI}: Lucy Edit is a text-based video editing model built on Wan 2.2~\cite{Hleihil2025DecartAI}, which directly takes an input video and a text instruction to produce an edited output.

\vspace{-5pt}
\subsection{Qualitative Results}
\vspace{-3pt}

\begin{figure}[t!]
    \centering
    \includegraphics[width=\linewidth]{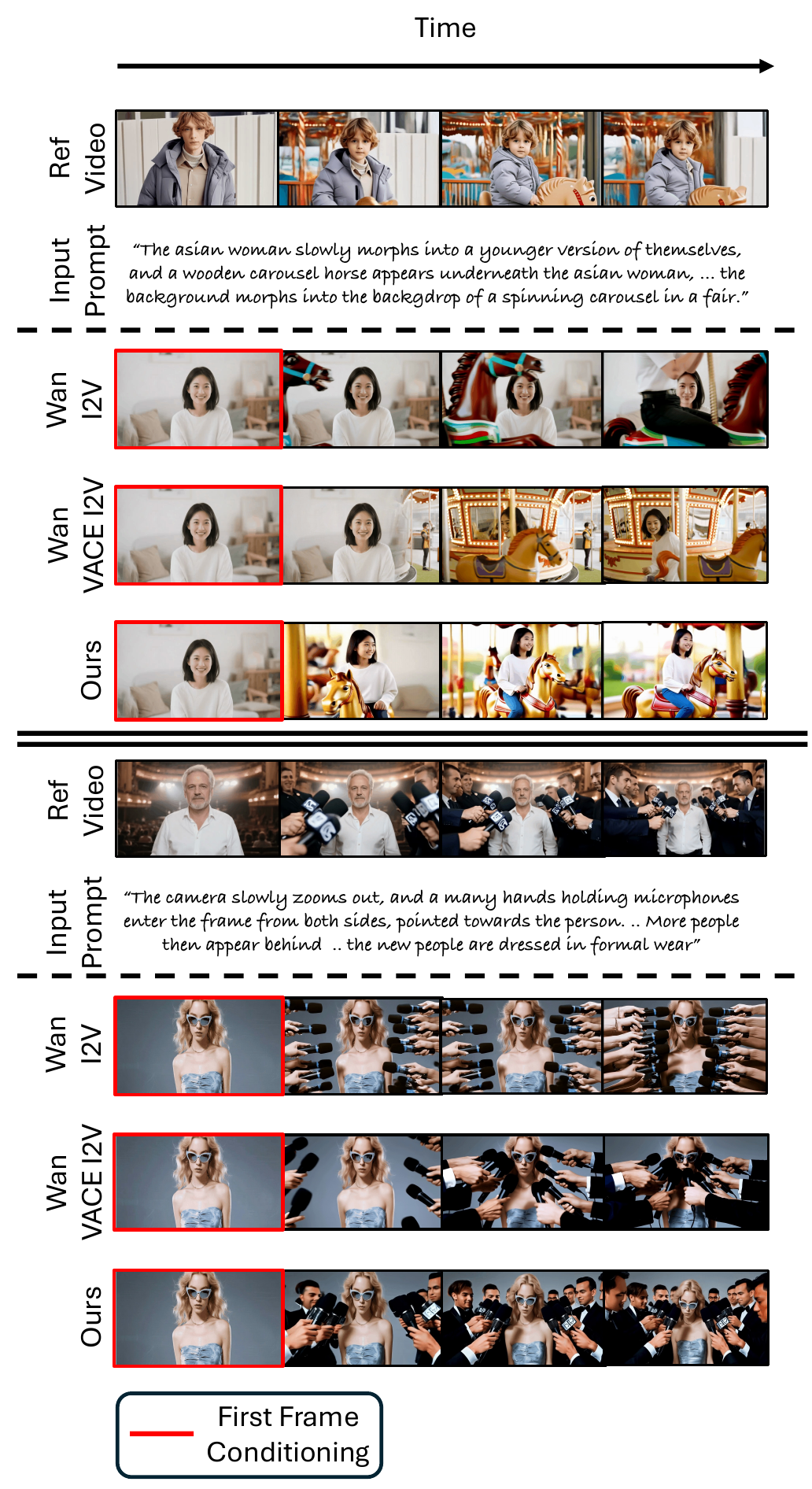}
        \vspace{-15pt}
    \caption{\textbf{Qualitative Reference Based Image to Video Results.} In the top example, baselines are unable to turn the input woman into a younger version of herself, or put her on the carousel. Our method correctly makes the subject younger, puts her on the carousel, and mimics the camera motion of the reference video. In the bottom example, baselines are unable to add the reporters into the scene, while our method correctly adds hands with microphones followed by reporters, and mimics the interactions and occlusions from the reference video.}
    \label{fig:results_i2v}
    \vspace{-20pt}
\end{figure}

\myparagraph{Image-to-Video Effects.}
Figure~\ref{fig:results_i2v} shows qualitative comparisons for the image-to-video task using unseen reference effects. Baseline models, which rely solely on text prompts, struggle to capture fine-grained temporal and stylistic cues. In contrast, our method accurately transfers the reference video’s camera motion, lighting, and character transformations, producing coherent temporal dynamics that align closely with the target effect.

\begin{figure*}[t!]
    \centering
    \includegraphics[width=\linewidth]{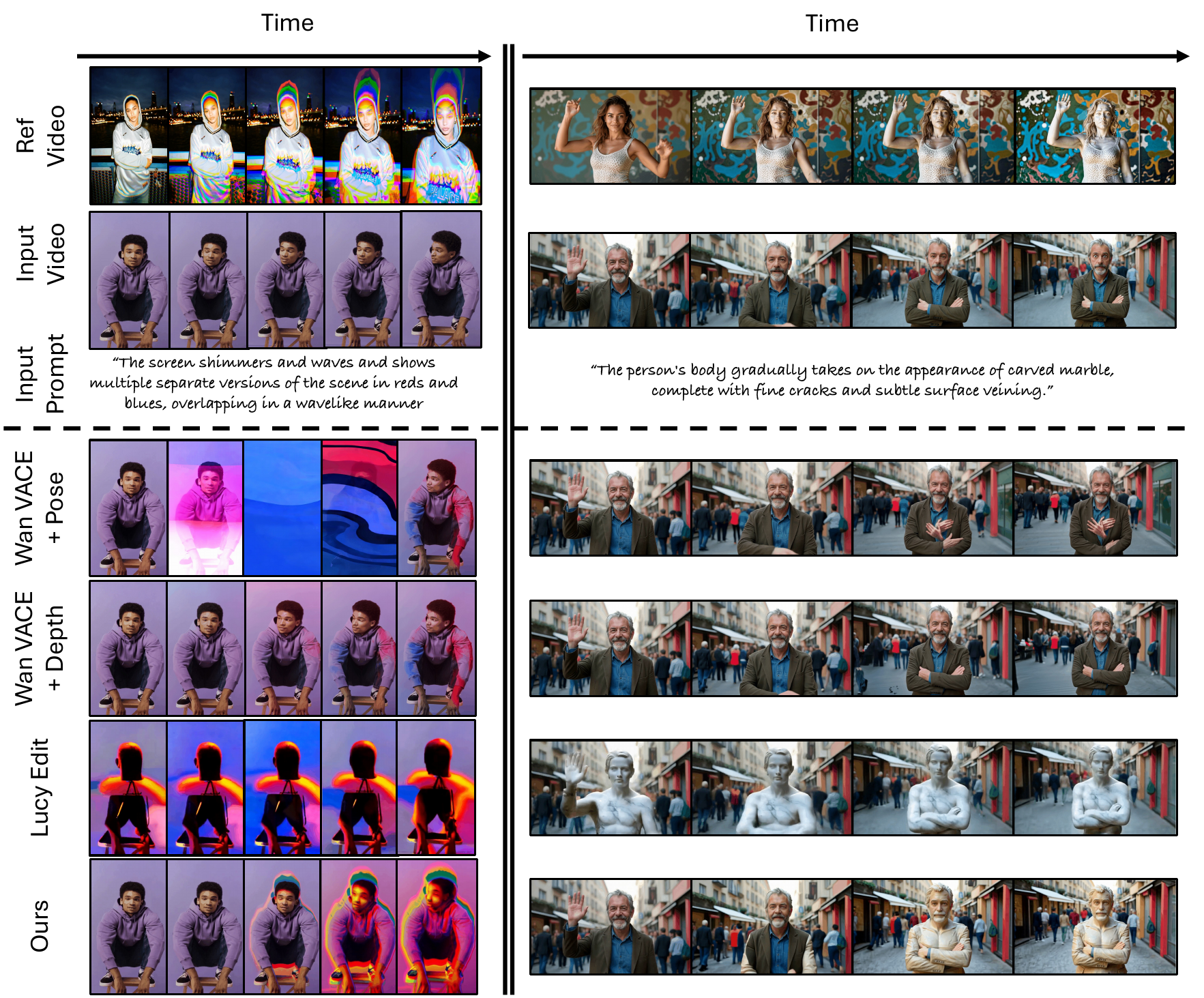}
        \vspace{-5pt}
    \caption{\textbf{Results for our reference-based video-to-video setup}. We compare three baselines: (1) Wan VACE + Pose: a conditional diffusion transformer \cite{jiang2025vace} using the first input frame and extracted poses; (2) Wan VACE + Depth: using the first frame and extracted depths with Wan VACE; and (3) Lucy Edit \cite{Hleihil2025DecartAI}, a video editing model fine-tuned from an image-to-video model. Without reference video information, all baselines fail to follow the reference from a prompt alone. Wan VACE methods overfit to the input or add irrelevant edits, while Lucy Edit applies uniform static edits across frames. Our method successfully follows the reference effect.}
    \label{fig:results_v2v}
    \vspace{-13pt}
\end{figure*}

\myparagraph{Video-to-Video Effects.}
Qualitative comparisons for the video-to-video task are shown in Figure~\ref{fig:results_v2v}. Wan VACE variants either overfit to the input video or introduce artifacts, while Lucy Edit produces static, frame-invariant edits that lack temporal evolution. By conditioning on the full reference video, our method consistently reproduces complex time-varying transformations such as gradual color shifts or structural morphing while maintaining spatial consistency and motion alignment with the input.

\begin{figure*}[t!]
    \centering
    \includegraphics[width=.9\linewidth]{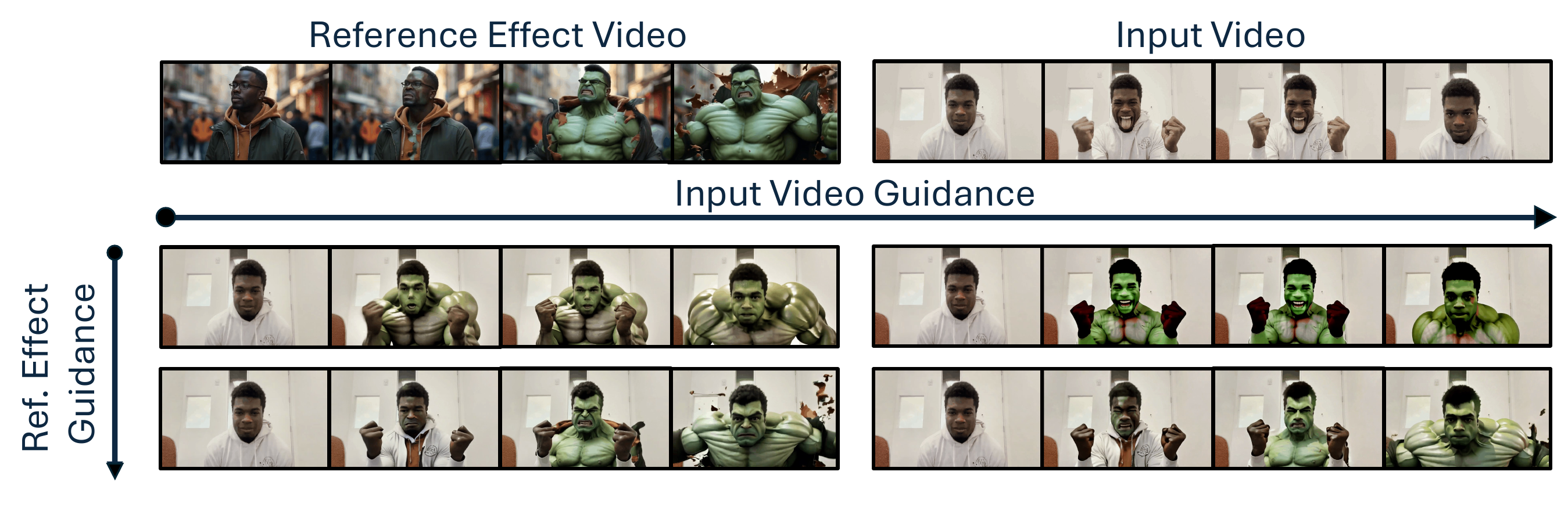}
        \vspace{-10pt}
    \caption{\textbf{Qualitative Results for Controllable Real Video Editing.} We show an example of the effect of reference effect guidance and input video guidance on model output. The first row shows the reference effect video and input video that are used for model conditioning. From left to right, we increase the input video guidance, while from top to bottom we increase the reference effect guidance. Notice that as input video guidance increases, the input video details are more faithfully kept (i.e. attire, facial shape). As the reference effect guidance increases, the reference effect video takes a larger role (i.e. reference effect muscle shape, and clothing of the man in the reference effect). Applying both gives a good mix of both input video adherence while following the effect. }
    \label{fig:ref_in_interpolation}
    \vspace{-15pt}
\end{figure*}

\myparagraph{User Study}
For a subjective task such as reference-based video editing or image-to-video generation, it is difficult to determine any single quantitative metric that fully captures success. We therefore conduct a comprehensive user study as the primary evaluation of our method across three tasks: neural video-to-video effect transfer (V2V), image-to-video effect transfer (I2V), and code-based temporal effects. Annotators on Amazon Mechanical Turk performed two-alternative forced choice (2AFC) pairwise comparisons between our method and several baselines, including Lucy Edit, VACE (Depth/Pose/I2V variants), Wan2.1, and an ablation that drops reference effect video conditioning (No Ref). Each comparison was evaluated along three dimensions: Reference Video Adherence (RVA), Input Video Adherence (IVA), and Overall Match (OM), where OM asks which output best applies the reference effect while preserving the input video. For I2V, we evaluate on RVA and OM. As shown in Table~\ref{tab:v2v_code_user_study_detailed}, our method achieves consistent preference over all baselines in V2V across all three dimensions, with win rates of 57.0--62.3\% in RVA, 53.7--60.7\% in IVA, and 60.0--67.2\% in OM. Code-based effects show even stronger results, with win rates of 59.5--65.7\% in RVA and 61.3--65.3\% in OM. For I2V (Table~\ref{tab:i2v_user_study_detailed}), we observe win rates of 56.8--57.4\% in RVA and 61.8--63.4\% in OM against all baselines. All reported win rates exceed the 50\% chance level, standard deviations confirm statistical reliability. The closest statistic in terms of preference is the input video adherence for our method without reference input video, which is expected as it uses the same model as our full method, and the test is simply adherence to the input video which both versions take as input. These results validate our approach's effectiveness in effect transfer while preserving input content fidelity across multiple generation modalities. \textbf{In total, we collected over 15000 annotations from over 500 unique annotators.} For more details on the user study setup, please refer to Section \ref{sec:more_user_study_details} of the Appendix.

\subsection{Quantitative Metrics}
\vspace{-2pt}
\myparagraph{Conditioning Input Similarity.}
Since our validation sets do not include ground-truth triplets, we measure the similarity of our generated outputs to both the input and reference videos as a proxy for task success. To quantify this, we employ VideoPrism~\cite{zhao2024videoprism}, a large-scale video embedding model pretrained on over 500M clips, to compute feature-space similarities between videos. For the reference + image-to-video case, we instead measure the average similarity between the generated video and its first-frame conditioning using CLIP~\cite{radford2021learning} embeddings. Results are summarized in Tables~\ref{tab:i2v_results_table} and~\ref{tab:v2v_results_table}.
Our method consistently achieves higher similarity to the reference video than all baselines, confirming that it effectively incorporates temporal and stylistic information from the reference. Interestingly, baseline models often exhibit slightly higher similarity to the input, likely reflecting a form of under-editing. This phenomenon is also visible in Figure~\ref{fig:results_v2v} (left) and Figure~\ref{fig:results_i2v} (bottom), where baselines preserve input structure but fail to reproduce the desired temporal evolution.

\vspace{-1 pt}

\subsection{Controllable Real Video Editing}
When editing real videos, users may want to control the level to which the original video is maintained, as well as the amount that the reference effect video effects the final generation. To this end, we follow previous work \cite{ho2022classifier, brooks2023instructpix2pix, jones2024customizing, kumari2025generating, kumari2024customizing, song2025history} and combine multiple classifier-free guidance directions during inference. Specifically, we consider applying guidance for text, input video, and reference effect video, normalizing latents as recomended in previous work \cite{kumari2025generating}. This gives: 
\vspace{-6pt}
\begin{equation}
\begin{split}
& v_\theta\!\left(x^t, x_{\text{ref}}, x_{\text{input}}, \varnothing \right)
+ \\ 
& \lambda_c \frac{\lVert g \rVert}{\lVert g_c \rVert} \, g_c
+ \lambda_{\text{ref}} \frac{\lVert g \rVert}{\lVert g_{\text{ref}} \rVert} \, g_{\text{ref}}
+ \lambda_{\text{in}} \frac{\lVert g \rVert}{\lVert g_{\text{in}} \rVert} \, g_{\text{in}},
\end{split}
\end{equation}

where
\begin{align}
g_c &= v_\theta\!\left(x^t, x_{\text{ref}}, x_{\text{input}}, c \right)
      - v_\theta\!\left(x^t, x_{\text{ref}}, x_{\text{input}}, \varnothing \right), \\
g_{\text{ref}} &= v_\theta\!\left(x^t, x_{\text{ref}}, x_{\text{input}}, c \right)
      - v_\theta\!\left(x^t, \varnothing, x_{\text{input}}, c \right), \\
g_{\text{in}} &= v_\theta\!\left(x^t, x_{\text{ref}}, x_{\text{input}}, c \right)
      - v_\theta\!\left(x^t, x_{\text{ref}}, \varnothing, c \right), \\
\lVert g \rVert &= \min\!\left( \lVert g_c \rVert, \lVert g_{\text{ref}} \rVert, \lVert g_{\text{in}} \rVert \right).
\end{align}
and $x^t$ is the noisy video latent, $x_{\text{ref}}$ is the clean reference video effect latent, $x_{\text{input}}$ is the clean input video latent, and $c$ is the text conditioning. Interpolating $\lambda_{\text{ref}}$ and $\lambda_{\text{in}}$ results in output videos with difference amounts of adherence to the reference effect video and input video respectively, as seen in Figure \ref{fig:ref_in_interpolation}.

\begin{table}[t]
\centering

\resizebox{.75\linewidth}{!}{
\begin{tabular}{lcc}
\toprule
\cmidrule(lr){2-3}
\textbf{Method} & First Frame Sim. & Ref Sim. \\
\midrule
Wan 2.1 & \textbf{0.7911} & \underline{0.7230} \\
Wan VACE I2V & \underline{0.7799} & 0.7127 \\
Ours & 0.7698 & \textbf{0.7378} \\
\bottomrule
\end{tabular}}
\caption{I2V Similarities. First Frame Sim is average clip \cite{radford2021learning} similarity to the input frame, and Ref Sim. is video embedding similarity \cite{zhao2024videoprism} between the output video and the reference effect video}
\label{tab:i2v_results_table}
\vspace{-10pt}
\end{table}
\begin{table}
\centering

\resizebox{.8\linewidth}{!}{
\begin{tabular}{lcccc}
\toprule
& \multicolumn{2}{c}{Neural V2V} & \multicolumn{2}{c}{Code Based V2V} \\
\cmidrule(lr){2-3}\cmidrule(lr){4-5}
\textbf{Method} & Input Sim. & Ref Sim. & Input Sim. & Ref Sim. \\
\midrule
Wan VACE Pose & \underline{0.9068} & 0.6539 & 0.9225 & \underline{0.6002} \\
Wan VACE Depth & \textbf{0.9460} & 0.6226 & \underline{0.9394} & 0.5998 \\
Lucy Edit & 0.7544 & \underline{0.6852} & 0.8882 & 0.6539 \\
Ours & 0.8568 & \textbf{0.7014} & \textbf{0.9479} & \textbf{0.7169} \\
\bottomrule
\end{tabular}}
\caption{Neural and Code-Based V2V Similarities. Input Sim. is video embedding similarity \cite{zhao2024videoprism} between the output video and the input video, and Ref Sim. is the video embedding similarity between the output video and the reference effect video.}
\label{tab:v2v_results_table}
\vspace{-15pt}
\end{table}

\section{Discussion}
\vspace{-5pt}
In conclusion, we presented \methodname{}, a tuning-free framework for transferring complex temporal visual effects across videos using reference conditioning. 
By introducing a large-scale dataset of over 120K triplets encompassing 1,700 distinct effects, we enable the first systematic study of reference-based temporal effect transfer. 
Our scalable video-to-video generation pipeline produces diverse, motion-consistent training pairs, while our diffusion-based architecture jointly encodes reference dynamics, input appearance, and semantic guidance to produce coherent results. 
Perceptual and quantitative evaluations demonstrate that \methodname{} surpasses prompt-only and static reference baselines in both visual fidelity and temporal coherence. 

\section{Acknowledgments}
\vspace{-5pt}
We would like to thank Yusuf Dalva, Rishubh Parihar, James Burgess, Ruihang Zhang, Sheng-Yu Wang,
Gaurav Parmar, and Nupur Kumari for
their insightful feedback and input that contributed to the
finished work. Maxwell Jones is
supported by the Rales Fellowship. This project is partly supported by the Packard Fellowship.

{
    \small
    \bibliographystyle{ieeenat_fullname}
    \bibliography{main}
}

\clearpage
\appendix
In Section \ref{sec:limitations}, we present limitations of our generated model, and show a qualitative result of a current limitation. In Section \ref{sec:dataset_details}, we provide detailed descriptions of our dataset creation process, including the generation pipelines and effect categorization. Section \ref{sec:more_user_study_details} goes over additional user study details, and Section \ref{sec:more_imp_details} outlines additional implementation details such as training configurations, inference settings, and computational requirements. In Section \ref{sec:more_quant_results}, we report extended quantitative comparisons between our method and baseline models across multiple metrics. 
\textbf{We provide extensive qualitative video results including outputs from our method, comparisons with baselines, and demonstrations from our new dataset in the accompanying website.html file located in the website folder of the supplement}.

\section{Limitations}
\label{sec:limitations}
\begin{figure}[t!]
    \centering
    \includegraphics[width=\linewidth]{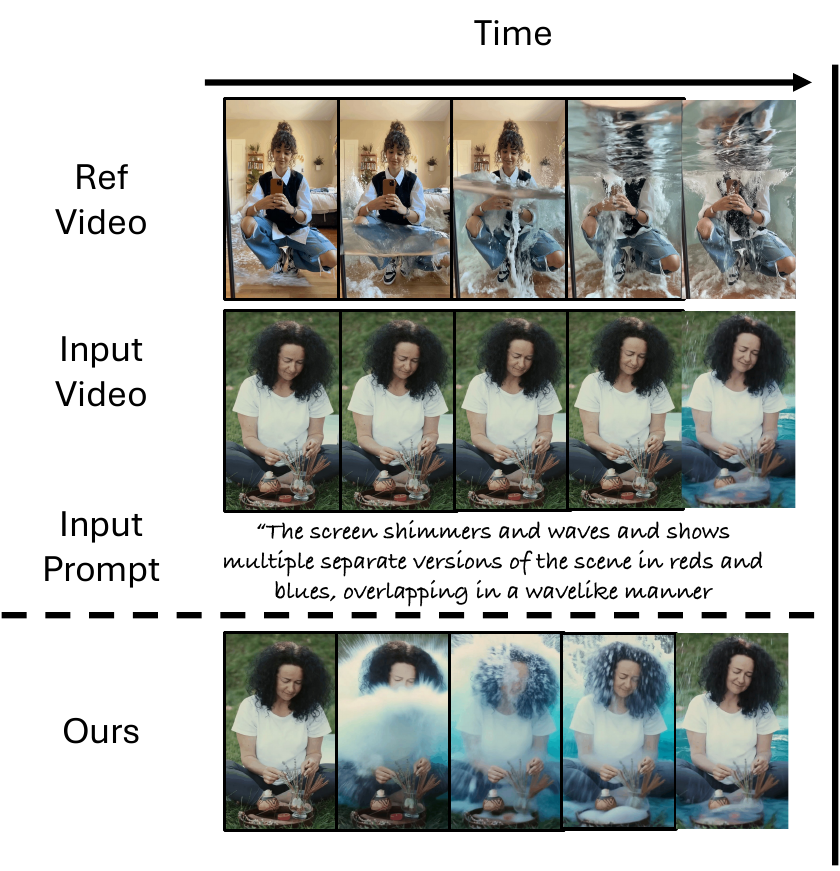}
        \vspace{-15pt}
    \caption{\textbf{Limitations.} We present a limitation of our model. Our model can struggle with imitating complete occlusion of parts of the body. In this example, the reference video has their entire head occluded by the rising water. Further, our method mistakenly adds a splashing effect that temporarily covers the subject but then subsides, as opposed to submerging them in standing water}
    \label{fig:limitations}
    \vspace{-20pt}
\end{figure}

While RefVFX achieves strong temporal coherence and visual fidelity across diverse effects, it still faces certain limitations. First, the model can struggle with accurately reproducing fine-grained occlusions or complex interactions between subjects and dynamic effects, occasionally leading to partial blending or misalignment artifacts. Second, our dataset primarily focuses on human-centric and foreground-dominant scenes, which may limit generalization to large-scale environmental effects or abstract cinematic transformations. We show an example of such limitations in Figure \ref{fig:limitations}. Inference remains computationally expensive due to the dual conditioning on both input and reference videos, resulting in roughly twice the generation time compared to single-source baselines. Addressing these limitations presents promising directions for future work.

\section{Dataset Details}
\label{sec:dataset_details}
We source the Image-to-Video data from open-source LoRA effects trained on top of the Wan 2.1 Image-to-Video model \cite{wan2025wan}, available on Hugging Face \cite{WanI2VLoRAs}. A comprehensive list of LoRA effects and their corresponding captions is provided in Table \ref{tab:lora_prompts}.

We use Qwen Image \cite{wu2025qwen} and Qwen Image Edit \cite{wu2025qwen}, together with Wan 2.1 VACE-1.3B \cite{wan2025wan}, to implement our text-guided video editing pair creation pipeline, as described in Figure 4 of the main text. We categorize our effect types and general effect categories in Table \ref{tab:effect_catalog_compact}.

For the code-based effects, we source data from the grounding subset of the Senorita dataset \cite{zi2025se}. Specifically, we filter for human-centric videos and include only grounding results with human-centric masks, yielding a collection of over 100K videos. The set of objects considered is listed in Table \ref{tab:senorita_grounding_objects}. In the original dataset, input videos are used to predict grounding masks; we repurpose these masks to obtain foreground or object and background segmentations, which we use to augment our effect pipelines. We manually curate code across a diverse set of effects and temporal transitions to sample from when creating triplets of the form (reference video, input video, output video). The full set of effect types and their corresponding hyperparameters is provided in Table \ref{tab:code_effects_hparams}, and the full set of temporal effect types and their corresponding hyperparameters is given in Table \ref{tab:temporal_effects_hparams}.

\section{User Study Details}
\label{sec:more_user_study_details}
\begin{figure*}[t!]
    \centering
    \includegraphics[width=\linewidth]{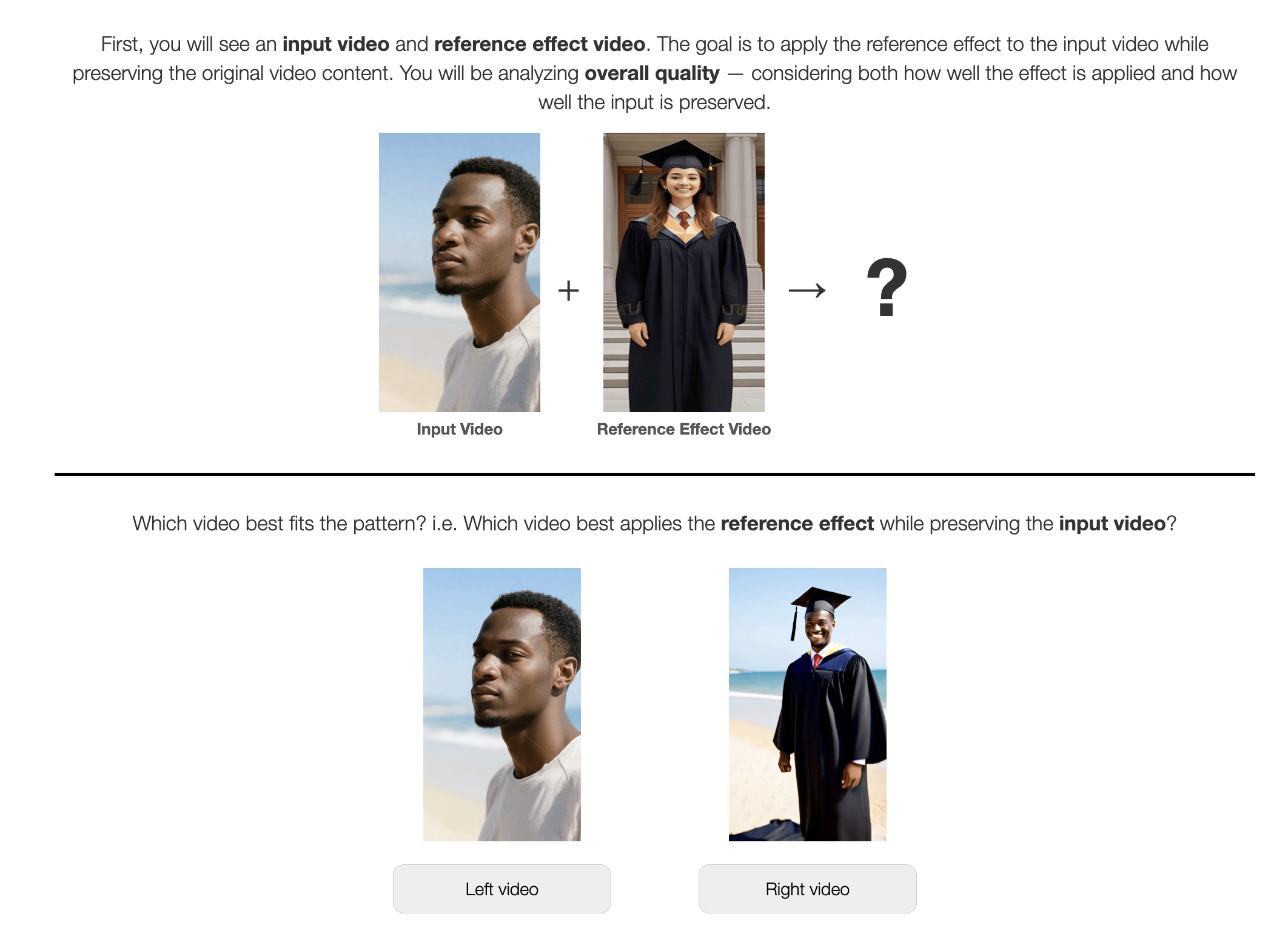}
        \vspace{-15pt}
    \caption{\textbf{User Study Example.} Example of a task shown to users. Users are prompted with the input video and reference video, and asked to pick which of the videos shown are better with resepct to either input video adherence, reference video adherence, or overall match to the task.}
    \label{fig:user_study}
    \vspace{-20pt}
\end{figure*}

We conduct our user study on Amazon Mechanical Turk (MTurk) using a two-alternative forced choice (2AFC) paradigm. In each trial, annotators are shown an input video and a reference effect video, followed by two output videos (one from our method and one from a baseline) presented side-by-side in randomized left/right order. Annotators select which output better satisfies the evaluation criterion. We evaluate three criteria in separate experiments: (1) \textbf{Reference Video Adherence (RVA)}: which output better applies the reference effect; (2) \textbf{Input Video Adherence (IVA)}: which output better preserves the input video; and (3) \textbf{Overall Match (OM)}: which output best applies the reference effect while preserving the input video.
Each experiment compares our method against a single baseline on a single evaluation dimension. Each experiment consists of 30 HITs (Human Intelligence Tasks), where each HIT is completed by a unique annotator. Within each HIT, annotators evaluate a subset of all available video pairs for that experiment, with the presentation order and left/right assignment randomized per HIT to mitigate ordering and side biases.
To ensure annotation quality, each HIT begins with 5 practice trials with immediate feedback. In practice trials, annotators are shown a clearly correct and clearly incorrect output and must select the correct one; feedback is provided after each practice trial to calibrate annotators before they proceed to the main evaluation.
We report win rates as the percentage of trials in which annotators preferred our method, where 50 percent represents chance performance. To estimate uncertainty, we perform bootstrap resampling with 10{,}000 iterations, resampling both annotators and trials with replacement. We report the standard deviation of the bootstrap distribution as our measure of statistical variability. Across all experiments, our method achieves win rates consistently above 50 percent, with bootstrap standard deviations typically in the range of 2--5 percent, confirming the statistical reliability of the observed preferences. See the setup in Figure \ref{fig:user_study}.

\section{More Implementation Details}
\label{sec:more_imp_details}
We train our model on videos consisting of 33 frames at 15 frames per second, using a resolution of 480p for all training and inference experiments. During inference, we use a classifier-free guidance scale of 5 and perform 50 sampling steps per generation. Because our model conditions on both the input and reference effect videos, it requires approximately twice the number of latent tokens compared to the baseline model that omits reference conditioning. As a result, inference time is roughly doubled: baseline video generation takes about 3.5 minutes per video, whereas our method requires approximately 7 minutes per video on a single NVIDIA A100-SXM4 GPU.

\section{More Quantitative Results}
\label{sec:more_quant_results}
We further assess video fidelity and perceptual quality using established metrics from VBench~\cite{huang2024vbench}, including motion smoothness~\cite{wan2025wan}, aesthetic score~\cite{LAION-AI2022aesthetic-predictor}, and dynamic degree~\cite{teed2020raft, huang2024vbench}.

Motion quality measures temporal consistency by computing the average cosine similarity \cite{radford2021learning} between consecutive frames:
\begin{equation}
\sum_{i = 2}^N \text{cosine sim}(\text{CLIP}(f_i), \text{CLIP}(f_{i-1})).
\end{equation}
Higher values indicate smoother and more coherent motion.

Aesthetic score follows the setup in VBench \cite{huang2024vbench}, using CLIP embeddings passed through a linear layer trained to predict human-rated aesthetic quality from a large-scale dataset of labeled images \cite{schuhmann2022laion}.

Dynamic degree quantifies how often videos exhibit meaningful motion. For each video, we compute optical flow using RAFT \cite{teed2020raft} and identify the top 5 percent of pixels with the highest flow magnitude per frame. Frames exceeding a motion threshold are labeled as dynamic; a video is then classified as dynamic if a sufficient proportion of its frames meet this criterion. The dynamic degree of a group of videos is defined as the percentage of videos that are labeled dynamic. 
We show extra qualitative results in Table \ref{tab:single_stats}. Across these quantitative measures, all methods perform comparably, indicating that our method can successfully transfer reference effects without sacrificing overall video quality.

\begin{table*}[t]
\vspace{-20pt}
\centering
\begin{tabular}{p{0.18\textwidth} p{0.75\textwidth}}
\toprule
\textbf{LoRA Name} & \textbf{Text Prompts} \\
\midrule
Squish & In the video, a miniature \{subject\} is presented. The \{subject\} is held in a person's hands. The person then presses on the \{subject\}, causing a sq41sh squish effect. The person keeps pressing down on the \{subject\}, further showing the sq41sh squish effect. \\
\midrule
Rotate & The video shows a \{subject\}. The \{subject\} performs a r0t4tion 360 degrees rotation. \\
\midrule
Inflate & (1) The video shows a \{subject\}, then infl4t3 inflates it, form expanding like a whimsical, cartoonish balloon. (2) The video shows a \{subject\}, then infl4t3 inflates it, form expanding into a perfect, inflated sphere. \\
\midrule
Cakeify & The video opens on a \{subject\}. A knife, held by a hand, hovers over the \{subject\}. The knife then begins cutting into the \{subject\} to c4k3 cakeify it. As it slices, the inside of the \{subject\} is revealed to be cake with chocolate layers. \\
\midrule
Deflate & The video opens with a \{subject\}. As the video progresses, the \{subject\} begins to d3d1at3 deflate it, gradually shrinking and losing shape, eventually flattening completely into a lifeless, deflated mass. \\
\midrule
Crush & The video begins with a \{subject\}. A hydraulic press descends toward it. Upon contact, the press c5us4 crushes the \{subject\}, flattening and deforming it until it collapses inward. \\
\midrule
Muscle & \{subject\} smiles slightly, then t2k1s takes off clothes revealing a lean muscular body and shows off muscles, giving a friendly smile. \\
\midrule
Bride & The video begins with a \{subject\}, then the 8r1d3 bride effect occurs. The \{subject\} is now in a white wedding dress, holding a bouquet, with a sunny beige background. \\
\midrule
Puppy & The video begins with a close-up portrait of a \{subject\}. The background changes and then the pu11y puppy effect begins. The \{subject\} is now surrounded by puppies and pets them. \\
\midrule
Baby & The video starts with a \{subject\}. Then the image shifts to the 848y baby effect, with \{subject\} in front of a crib, surrounded by toys, then shown again in a different location as a baby version. \\
\midrule
VIP & The video begins with an image of \{subject\}. Then the v1p red carpet transformation appears — \{subject\} in a black dress, gold jewelry, photographed on the red carpet. \\
\midrule
Mona-Lisa & The video starts with an image of \{subject\}. The m0n4 Mona Lisa transformation begins, wrapping around the \{subject\}, who is then depicted as a Mona Lisa version seated before a landscape. \\
\midrule
Princess & The video begins with a \{subject\}. A pr1nc355 princess transformation occurs: sparkling light appears, and \{subject\} is now in a silver beaded gown with tiara and gloves, seated among gifts and candles. \\
\midrule
Pirate-Captain & The video begins with a \{subject\}. The p1r4t3 pirate captain transformation follows. The \{subject\} now wears a black hat, coat, and sash aboard a wooden ship with a sword. \\
\midrule
Samurai & The video begins with a \{subject\}. The 54mur41 samurai transformation turns them into a samurai wearing traditional armor with a katana, against a misty mountain backdrop. \\
\midrule
Zen & The video starts with a portrait of a \{subject\}. The z3n1fy zen transformation follows: pink robe, zen garden, and later a black kimono in a tranquil garden setting. \\
\midrule
Assasin & The video starts with a portrait of a \{subject\}. The 3p1c epic transformation begins. The \{subject\} wears a red coat, white hair, and black gloves, holding guns in both hands. \\
\midrule
Painting & The video starts with a \{subject\}. They appear in a gold framed mirror, then transform into a p41nt1ng painting version in red and blue attire with an old-style painted background. \\
\midrule
Disney-Princess & The video starts with a \{subject\}. The d15n3y princess transformation occurs: the \{subject\} is now in a blue dress, with butterflies falling in a classic hallway setting. \\
\midrule
Snow-White & The video begins with a \{subject\} outdoors, then cuts to the sn0w\_wh1t3 transformation: classic dress, red apple, forest background. \\
\midrule
Classy & The video starts with a \{subject\} in a suit. The c1455y transformation occurs — now in a light blue dress, smiling at the camera, seated with a flower and envelope. \\
\bottomrule
\end{tabular}
\caption{Subset of list of LoRA models and their corresponding text prompts used in this work. Each LoRA entry spans multiple lines for readability.}
\label{tab:lora_prompts}
\end{table*}

\begin{table*}[t]
\vspace{-25pt}
\centering
\begin{tabularx}{\textwidth}{>{\raggedright\arraybackslash}p{0.26\textwidth} Y}
\toprule
\textbf{Category} & \textbf{Paraphrased Effects} \\
\midrule

\textbf{Object Addition} &
Balance a tall stack of encyclopedias on head, wear a beekeeper suit with neon honeycomb glow, add a pizza cape, perch a tiny green dragon on shoulder, materialize toga + sandals + holographic smartwatch, hat becomes a live octopus, juggle three bright-green rubber chickens, use a banana as a phone, parade of wind-up toy robots follows behind, attach a leash to a pet cloud, grow a gnome hat and long white beard, put on a diver’s helmet (library setting), don an astronaut helmet + 19th-century ball gown, tuxedo T-shirt appears and tiny penguins roam shoulders, hands become giant foam novelty hands, endless silk scarves stream from ear, open a shimmering interdimensional portal, person covered head-to-toe in rainbow alien slime, business suit catches fire (person stays calm), three personal mini-planets orbit the person. \\

\midrule
\textbf{Weather + Atmospheric Effects} &
Morning fog rolls in and softens background, thick industrial smog mutes colors, gentle misty drizzle begins, heavy rainfall drenches the scene, post-rain surfaces glisten like mirrors, light picturesque snowfall, heavy snowstorm with blowing flakes, full-blown blizzard white-out, hailstones streak and bounce, sandstorm tints scene orange and dusty, smoke fills air as if from nearby fire, floating embers and rising ash, ash and embers fall from sky, visible dust motes in sunbeams, aurora borealis bathes scene in eerie glow, electric storm with flashing lightning, heat haze shimmers and warps background, underwater environment with flowing caustics, moonrise casts pale long shadows, volumetric light rays through window/trees. \\

\midrule
\textbf{Artistic \& Stylistic Effects} &
Risograph print look (overlapping magenta/cyan layers), stained-glass pane refractions, cross-hatched pencil sketch rendering, duotone (magenta + black), thermal-camera false-color map, infrared look (white foliage, dark sky), holographic scanline overlay, datamosh/pixel-sorting glitch, digital code “Matrix” rain, stop-motion claymation style, low-poly geometric rendering, blueprint schematic overlay, graphic-novel/comic-book inks, chalk drawing on dusty blackboard, aged fresco on cracked plaster, impressionistic oil painting, soft watercolor rendering, vintage 1970s film grain and fade, classic 35mm black-and-white film, sepia old-photo vignette, chrome-sphere reflection world, anamorphic lens flare, film-noir high-contrast lighting. \\

\midrule
\textbf{Particle \& Element Effects} &
Confetti shower from above, dandelion seeds drift by, white feathers fall like snow, rainbow-hued mist swirls, fireflies blink in dark areas, countless soap bubbles float, golden coins rain down, glitter rains from halo overhead, golden sparkles twinkle throughout, cherry blossom petals whirl, glowing monarch butterflies swarm, pixie-dust trails swirl magically. \\

\midrule
\textbf{Color Palette \& Tonal Changes} &
Deep monochrome blue grade, oversaturated vaporwave palette, selective color (B\&W except vivid red), radioactive sickly-green tint, sun-bleached faded desert look, autumnal warm grade (oranges/reds/browns), teal-and-orange blockbuster grade, high-key bright airy palette, cool wintry grade (blues/whites/grays), twilight blue after sunset, warm golden-hour glow, soft morning side-light, fluorescent clinical lighting, candlelight illumination, bonfire-lit scene with flicker, scene lit by bioluminescence, cyberpunk neon glow with digital rain. \\

\midrule
\textbf{Surreal \& Fantasy Transformations} &
Floating green-topped islands fill the sky, giant translucent spirit animals in background, overgrown with glowing bioluminescent fungi, two-sun alien planet landscape, holographic projection aura overlays scene, world fractures with spreading glass-like cracks, everything but person crystallizes, environment melts like a Dalí painting, background becomes swirling galaxy nebula, candy-themed fantasy world outside, impossible Escher-like architecture, gravity inverts and small objects float upward, person levitates into meditation pose, ground becomes still mirror-water, colors invert to negative film, ethereal mid-ground mist drifts through, shadow detaches and dances playfully, sticky notes stack up with bad cat doodles, psychedelic rainbow color-shifting sweep, psychedelic swirling liquid-paint background, scene inside a giant snow globe, neon city-night transition (cyberpunk), gritty graphic stylization with bold outlines. \\

\bottomrule
\end{tabularx}
\caption{Grouped catalog of video effects. We provide a set of general categories for our video to video effects as described in Figure 4 of the main text. We provide a paraphrased version of each effect, as the actual prompts are long.
.}
\label{tab:effect_catalog_compact}
\end{table*}

\begin{table*}[t]

\centering
\begin{tabularx}{\textwidth}{>{\raggedright\arraybackslash}X}
\toprule
\textbf{Grounding Objects} \\
\midrule
a a woman, a person, a woman, baby, bookshelf, person, child, children, delivery person, eye, eye mask, face, feet, foot, glasses, hair, hand, hands, head, man, nose ring, old man, old woman, pathway person, pathway, person, people, person, person man, person person, person woman, pregnant woman, skateboard, person, the person, watch, person, woman, women, young man, young people, young woman. \\
\bottomrule
\end{tabularx}
\caption{List of grounding objects kept from Senorita dataset grounding subset.}
\label{tab:senorita_grounding_objects}
\end{table*}

\begin{table*}[t]

\centering
\begin{tabularx}{\textwidth}{>{\raggedright\arraybackslash}p{0.33\textwidth} Y}
\toprule
\textbf{Effect} & \textbf{Hyperparameters} \\
\midrule
Posterize Frames & color palette (30 options; 4/6/8 colors each), buckets = palette size \\
Pixelate Frames & pixel length: \{4, 8, 16, 32\} \\
Invert Frames & (no hyperparameters) \\
Wave Warp & amplitude: [10, 50], frequency: [0.01, 0.075] \\
Update Saturation Brightness & brightness: [-255, 254], saturation: \{0.0, 0.5, 1.0, 1.5, 2.0\} \\
Gaussian Blur & kernel size: \{(11,11), (21,21), (51,51), (101,101), (151,151)\} \\
Add Grain & amount: [10, 50], grain size: \{1, 2, 3, 4, 5\} \\
Black And White & (no hyperparameters) \\
Color Overlay & percent overlay: \{0.25, 0.5, 0.75, 1.0\}, color (RGB 0–255 each) \\
CC Ball Action & grid spacing: \{5, 10, 20, 30\}, ball color (RGB 0–255 each) \\
Sticker Effect & border size: \{10, 20, 30, 40, 50\}, sticker color (RGB 0–255 each) \\
Glow Effect & glow size: \{10, 20, 30, 40, 50\}, glow color (RGB 0–255 each), glow brightness: \{0.5, 1.0, 1.5, 2.0\}, object brightness: \{0, 1, 2, 3, 4, 5\} \\
Radial Blur & center: (0.5, 0.5) or uniform [0.10, 0.90] (2dp), strength: [5, 60], blur border (\% of corner distance): [0, 30] \\
Rotate Pixels & center: uniform [0.000, 1.000] (3dp), max angle: \{10, 20, 30, 40, 50\}, radius (px): \{100, 200, 300, 400, 500\} \\
Glitch Effect & angle: [0, 359], red displacement: [-40, 40], green displacement: [-40, 40], blue displacement: [-5, 5] \\
Dither & size: \{5, 8, 10, 12, 16, 20\}, color steps per channel: \{2, 3, 4, 5, 7, 9\} \\
Motion Blur & angle: [0, 360], strength: [0.1, 0.6] \\
Stutter & hold duration (frames): \{1, 2\}, stutter frequency (every Nth frame): \{3, 4, 5, 6\} \\
Ghosting & ghost intensity: [0.1, 0.9] \\
Strobe & flash frequency (frames): \{4, 6, 8, 10\}, flash duration (frames): \{1, 2\}, flash color: \{white, black\} \\
\bottomrule
\end{tabularx}
\caption{Code-based video effects and their hyperparameters. Ranges indicate the values sampled or supported by the provided config generators.}
\label{tab:code_effects_hparams}
\end{table*}

\begin{table*}[t]

\centering
\begin{tabularx}{\textwidth}{>{\raggedright\arraybackslash}p{0.40\textwidth} Y}
\toprule
\textbf{Temporal Effect} & \textbf{Hyperparameters} \\
\midrule
Alpha Blend & softness: [0.01, 0.05], center: (0.5, 0.5) or uniform [0.00, 1.00] (2dp), transition window (frames): start and end in [0, 33] \\
Wipe Left To Right & softness: [0.01, 0.05], center: (ignored), transition window (frames): start and end in [0, 33] \\
Wipe Right To Left & softness: [0.01, 0.05], center: (ignored), transition window (frames): start and end in [0, 33] \\
Wipe Top To Bottom & softness: [0.01, 0.05], center: (ignored), transition window (frames): start and end in [0, 33] \\
Wipe Bottom To Top & softness: [0.01, 0.05], center: (ignored), transition window (frames): start and end in [0, 33] \\
Diag Top Left Bottom Right & softness: [0.01, 0.05], center: (ignored), transition window (frames): start and end in [0, 33] \\
Diag Bottom Right Top Left & softness: [0.01, 0.05], center: (ignored), transition window (frames): start and end in [0, 33] \\
Diag Top Right Bottom Left & softness: [0.01, 0.05], center: (ignored), transition window (frames): start and end in [0, 33] \\
Diag Bottom Left Top Right & softness: [0.01, 0.05], center: (ignored), transition window (frames): start and end in [0, 33] \\
Circle Out & softness: [0.01, 0.05], center: (0.5, 0.5) or uniform [0.00, 1.00] (2dp), transition window (frames): start and end in [0, 33] \\
Circle In & softness: [0.01, 0.05], center: (0.5, 0.5) or uniform [0.00, 1.00] (2dp), transition window (frames): start and end in [0, 33] \\
Rect Out & softness: [0.01, 0.05], center: (0.5, 0.5) or uniform [0.00, 1.00] (2dp), transition window (frames): start and end in [0, 33] \\
Rect In & softness: [0.01, 0.05], center: (0.5, 0.5) or uniform [0.00, 1.00] (2dp), transition window (frames): start and end in [0, 33] \\
Diamond Out & softness: [0.01, 0.05], center: (0.5, 0.5) or uniform [0.00, 1.00] (2dp), transition window (frames): start and end in [0, 33] \\
Diamond In & softness: [0.01, 0.05], center: (0.5, 0.5) or uniform [0.00, 1.00] (2dp), transition window (frames): start and end in [0, 33] \\
\bottomrule
\end{tabularx}
\caption{Temporal transition effects and their hyperparameters. Note that start frame $<$ end frame always when deciding transition window}
\label{tab:temporal_effects_hparams}
\end{table*}

\begin{table*}

\begin{tabular}{lccc ccc ccc}
\toprule
& \multicolumn{3}{c}{I2V} & \multicolumn{3}{c}{Neural V2V} & \multicolumn{3}{c}{Code Based V2V} \\
\cmidrule(lr){2-4}\cmidrule(lr){5-7}\cmidrule(lr){8-10}
\textbf{Method} & AES & Motion & Dynamic & AES & Motion & Dynamic & AES & Motion & Dynamic \\
\midrule
Ours & 0.5607 & 0.975 & 0.711 & 0.5649 & 0.969 & 0.712 & 0.4802 & 0.984 & 0.143 \\
Wan 2.1 & 0.5559 & 0.977 & 0.737 & - & - & - & - & - & - \\
Wan VACE I2V & 0.5479 & 0.974 & 0.816 & - & - & - & - & - & - \\
Wan VACE Pose & - & - & - & 0.5549 & 0.968 & 0.817 & 0.5270 & 0.992 & 0.190 \\
Wan VACE Depth & - & - & - & 0.5977 & 0.975 & 0.421 & 0.5532 & 0.994 & 0.048 \\
Lucy Edit & - & - & - & 0.5075 & 0.987 & 0.150 & - & - & - \\
Validation & 0.5557 & 0.972 & 0.765 & 0.6435 & 0.976 & 0.750 & 0.4922 & 0.985 & 0.190 \\
\bottomrule
\end{tabular}
\caption{Individual video statistics across methods.}
\label{tab:single_stats}
\end{table*}

\end{document}